\documentclass[11pt,a4paper]{article}
\usepackage[hyperref]{naaclhlt2019}
\usepackage{times}
\usepackage{latexsym}
\usepackage{url}
\usepackage[english]{algorithm2e}
\usepackage{booktabs}
\usepackage{enumitem}
\usepackage{subfig}
\usepackage{graphicx}
\usepackage{multirow}
\usepackage{epsfig}
\aclfinalcopy 

\title{Posterior-regularized REINFORCE for Instance Selection in Distant Supervision}

\author{\textbf{Qi Zhang$^1$, Siliang Tang$^1$\thanks{Corresponding author}, Xiang Ren$^3$,Fei Wu$^1$, Shiliang Pu$^2$ \& Yueting Zhuang$^1$}\\
    $^1$Zhejiang University, 
    $^2$Hikvision, $^3$University of Southern California, \\
    \texttt{\{zhangqihit,siliang,wufei,yzhuang\}@zju.edu.cn},\\ \texttt{pushiliang@hikvision.com},\\ \texttt{xiangren@usc.edu}}

\date{}

\begin{document}

\maketitle
\begin{abstract}
This paper provides a new way to improve the efficiency of the REINFORCE training process. We apply it to the task of instance selection in distant supervision.  Modeling the instance selection in one bag as a sequential decision process, a reinforcement learning agent is trained to determine whether an instance is valuable or not and construct a new bag with less noisy instances. However unbiased methods, such as REINFORCE, could usually take much time to train. This paper adopts posterior regularization (PR) to integrate some domain-specific rules in instance selection using REINFORCE. As the experiment results show, this method remarkably improves the performance of the relation classifier trained on cleaned distant supervision dataset as well as the efficiency of the REINFORCE training.
\end{abstract}
\section{Introduction}
Relation extraction is a fundamental work in natural language processing. Detecting and classifying the relation between entity pairs from the unstructured document, it can support many other tasks such as question answering.

While relation extraction requires lots of labeled data and make methods labor intensive,  ~\cite{mintz2009} proposes distant supervision (DS), a widely used automatic annotating way. In distant supervision, knowledge base (KB) , such as Freebase, is aligned with nature documents. In this way, the sentences which contain an entity pair in KB all express the exact relation that the entity pair has in KB. We usually call the set of instances that contain the same entity pair a bag. In this way, the training instances can be divided into N bags $\textbf{B}=\{ B^1,B^2,...,B^N \}$. Each bag $B^k$ are corresponding to an unique entity pair $E^k=(e^k_{1},e^k_{2})$ and contains a sequence of instances $\{x^k_{1},x^k_{2},...,x^k_{|B^k|}\}$ . However, distant supervision may suffer a wrong label problem. In other words, the instances in one bag may not actually have the relation.

To resolve the wrong label problem, just like Fig.2 shows,  \cite{feng2018reinforcement} model the instance selection task in one bag $B^k$ as a sequential decision process and train an agent $\pi(a|s,\theta_{\pi})$ denoting the probability $P_{\pi}(A_{t}=a,|S_{t}=s,\theta_{t}=\theta_{\pi})$  that action $a$ is taken at time $t$ given that the agent is in state $s$ with parameter vector $\theta_{\pi}$ by REINFORCE algorithm ~\cite{bookrl}. The action $a$ can only be 0 or 1 indicating whether an instance $x^k_{i}$ is truly expressing the relation and whether it should be selected and added to the new bag $\overline{B^k}$. The state $s$ is determined by the entity pair corresponding to the bag, the candidate instance to be selected and the instances that have already been selected. Accomplishing this task, the agent gets a new bag $\overline{B^k}$ at the terminal of the trajectory with less wrong labeled instances. With the newly constructed dataset $\overline{\textbf{B}}=\{ \overline{B^1},\overline{B^2},...,\overline{B^N} \}$ with less wrong labeling instances, we can train bag level relation predicting models with better performance. Meanwhile, the relation predicting model gives reward to the instance selection agent. Therefore, the agent and the relation classifier can be trained jointly.

However, REINFORCE is a Monte Carlo algorithm and need stochastic gradient method to optimize. It is unbiased and has good convergence properties but also may be of high variance and slow to train~\cite{bookrl}.

\begin{figure}[bhp]
\includegraphics[height=80pt,width=\columnwidth]{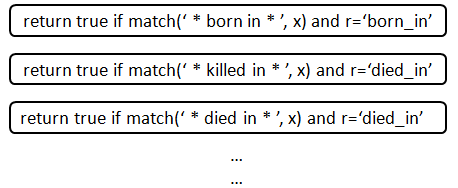}
\caption{Rule Pattern Examples }
\end{figure}

Therefore, we train a REINFORCE based agent by integrating some other domain-specific rules to accelerate the training process and guide the agent to explore more effectively and learn a better policy. Here we use a rule pattern as the Fig.1 shows~\cite{Liu2017Heterogeneous}. The instances that return true (match the pattern and label in any one of the rules) are denoted as $x_{MI}$ and we adopt posterior regularization method~\cite{PR2010} to regularize the posterior distribution of $\pi(a|s,\theta_{\pi})$ on $x_{MI}$. In this way, we can construct a rule-based agent $\pi_{r}$. $\pi_{r}$ tends to regard the instances in $x_{MI}$ valuable and select them without wasting time in trial-and-error exploring. The number of such rules is 134 altogether and can match nearly four percents of instances in the training data.

Our contributions include:
\begin{itemize}
\item We propose PR REINFORCE by integrating domain-specific rules to improve the performance of the original REINFORCE.
\item We apply the PR REINFORCE to the instance selection task for DS dataset to alleviate the wrong label problem in DS.
\end{itemize}

\section{RELATED WORK}
Among the previous studies in relation extraction, most of them are supervised methods that need a large amount of annotated data \cite{bach2007review}. Distant supervision is proposed to alleviate this problem by aligning plain text with Freebase. However, distant supervision inevitably suffers from the wrong label problem.

Some previous research has been done in handling noisy data in distant supervision. An expressed-at-least-once assumption is employed in \cite{mintz2009}: if two entities participated in a relation, at least one instance in the bag might express that relation. Many follow-up studies adopt this assumption and choose a most credible instance to represent the bag. \cite{aclnre2016,ji2017distant} employs the attention mechanism to put different attention weight on each sentence in one bag and assume each sentence is related to the relation but have a different correlation.

Another key issue for relation extraction is how to model the instance and extract features, \cite{cnn2014,pcnn2015,BGRU2016} adopts deep neural network including CNN and RNN, these methods perform better than conventional feature-based methods.

Reinforcement learning has been widely used in data selection and natural language processing. \cite{feng2018reinforcement} adopts REINFORCE in instance selection for distant supervision which is the basis of our work.

Posterior regularization \cite{PR2010} is a framework to handle the problem that a variety of tasks and domains require the creation of large problem-specific annotated data. This framework incorporates external problem-specific information and put a constraint on the posterior of the model. In this paper, we propose a rule-based REINFORCE based on this framework.

\section{Methodology}
In this section, we focus on the model details. Besides the interacting process of the relation classifier and the instance selector, we will introduce how to model the state, action, reward of the agent and how we add rules for the agent in training process. 

\begin{figure}
\includegraphics[height=180pt, width=\columnwidth]{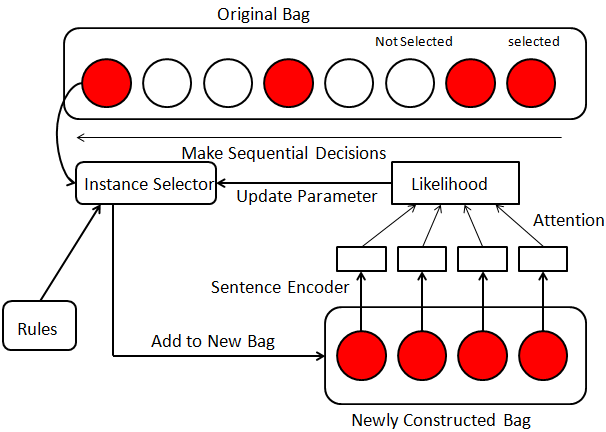}
\caption{Overall Framework}
\end{figure}

\subsection{basic relation classifier}

We need a pretrained basic relation classifier to define the reward and state. In this paper, we adopt the BGRU with attention  bag level relation classifier $f_{b}$ \cite{BGRU2016}. With \textbf{o} denoting the output of $f_{b}$ corresponding to the scores associated to each relation, the conditional probability can be written as follows:

\begin{equation}
 P_{f_{b}}(r|B^k,\theta_{b})= \frac{exp(o_{r})}{\sum_{k=1}^{n_{r}} exp(o_{k})}
\end{equation}
	
where $r$ is relation type, $n_{r}$ is the number of relation types, $\theta_{b}$ is the parameter vector of the basic relation classifier $f_{b}$ and $B^k$ denotes the input bag of the classifier.

In the basic classifier, the sentence representation is calculated by the sentence encoder network BGRU, the BGRU takes the instance $x^k_{i}$ as input and output the sentence representation BGRU($x^k_{i}$). And then the sentence level(ATT) attention will take  $\{BGRU(x^k_{1}),BGRU(x^k_{2}),...,BGRU(x^k_{|B^k|})\}$ as input and output  \textbf{o} which is the final output of $f_{b}$ corresponding to the scores associated to each relation.

\subsection{Original REINFORCE}
Original REINFORCE agent training process is quite similar to \cite{feng2018reinforcement}.
The instance selection process for one bag is completed in one trajectory. Agent $\pi(a|s,\theta_{\pi})$ is trained as an instance selector.

The key of the model is how to represent the state in every step and the reward at the terminal of the trajectory. We use the pretrained $f_{b}$ to address this key problem. The reward defined by the basic relation classifier is as follows:
\begin{equation}
  R=\log P_{f_{b}}(r^k|\overline{B^k},\theta_{b})
\end{equation}

  In which $r^k$ denotes the corresponding relation of ${B^k}$.

   The state $s$ mainly contained three parts: the representation of the candidate instance, the representation of the relation and the representation of the instances that have been selected.

The representation of the candidate instance are also defined by the basic relation classifier $f_{b}$. At time step t, we use BGRU($x^k_{t}$) to represent the candidate instance $x^k_{t}$ and the same for the selected instances. As for the embedding of relation, we use the entity embedding method introduced in TransE model \cite{bordes2013translating} which is trained on the Freebase triples that have been mentioned in the training and testing dataset, and the relation embedding $re_{k}$ will be computed by the difference of the entity embedding element-wise.

The policy $\pi$ with parameter $\theta_{\pi}=\{W,b\}$ is defined as follows:
\begin{equation}
  P_{\pi}(A_{t}|S_{t},\theta_{\pi}) = softmax(WS_{t}+b)
\end{equation}
With the model above, the parameter vector can be updated according to REINFORCE algorithm \cite{bookrl}.

\subsection{Posterior Regularized REINFORCE}

REINFORCE uses the complete return, which includes all future rewards up until the end of the trajectory. In this sense, all updates are made after the trajectory is completed \cite{bookrl}. These stochastic properties could make the training slow. Fortunately, we have some domain-specific rules that could help to train the agent and adopt posterior regularization framework to integrate these rules. The goal of this framework is to restrict the posterior of $\pi$. It can guide the agent towards desired behavior instead of wasting too much time in meaninglessly exploring.

Since we assume that the domain-specific rules have high credibility, we designed a  rule-based policy agent $\pi_{r}$ to emphasize their influences on $\pi$. The posterior constrains for $\pi$ is that the policy posterior for $x_{MI}$ is expected to be 1 which indicates that agent should select the $x_{MI}$. This expectation can be written as follows:
\begin{equation}
 E_{P_{\pi}}[\textbf{l}(A_{t}=1)]=1
\end{equation}

where \textbf{l} here is the indicator function. In order to transfer the rules into a new policy $\pi_{r}$, the KL divergence between the posterior of $\pi$ and $\pi_{r}$ should be minimized, this can be formally defined as

\begin{equation}
  min  KL(P_{\pi}(A_{t}|S_{t},\theta_{\pi})||P_{\pi_{r}}(A_{t}|S_{t},\theta_{\pi}))
\end{equation}

Optimizing the constrained convex problem defined by Eq.(4) and Eq.(5), we get a new policy $\pi_{r}$:

\begin{equation}
  P_{\pi_{r}}(A_{t}|S_{t},\theta_{\pi}) = \frac{P_{\pi}(A_{t}|S_{t},\theta_{\pi})exp(\textbf{l}(A_{t}=1)-1)}{Z}
\end{equation}

where Z is a normalization term.
\begin{displaymath}
  Z=\sum_{A_{t}=0}^{1} P_{\pi_{r}}(A_{t}|X,\theta_{\pi})exp(\textbf{l}(A_{t}=1)-1)
\end{displaymath}

Algorithm 1 formally define the overall framework of the rule-based data selection process.

\begin{algorithm}[t]
\SetAlgoLined
\KwData{Original DS Dataset: $\textbf{B}=\{ B^1,B^2,...,B^N \}$, Max Episode:M, Basic Relation Classifier:$f_{b}$, Step Size: $\alpha$ }
\KwResult{An Instance Selector}
initialization policy weight $\theta_{\pi}'=\theta_{\pi}$\;
initialization classifier weight $\theta_{b}'=\theta_{b}$\;
\For{episode m=1 to M}{
\For{$B^k$ in \textbf{B} }{
$B^k=\{x^k_{1},x^k_{2},...,x^k_{|B^k|}\},\overline{B^k}=\{\}$\;
\For{step i in $|B^k|$}{
construct $s_{i}$ by $\overline{B^k}, x^k_{i},re_{k}$\;
\eIf{$x^k_{i}\in x_{MI}$}{
construct $\pi_{r}$\;
sample action $A_{i}$ follow $\pi_{r}(a|s_{i},\theta_{\pi}')$\;
}
{
sample action $A_{i}$ follow $\pi(a|s_{i},\theta_{\pi}')$\;
}
\If{$A_{i}$=1}{
 Add $x^k_{i}$ in $\overline{B^k}$\;
}
}
Get terminal reward: $R=\log P_{f_{b}}(r^k|\overline{B^k},\theta_{b}')$\;
Get step delayed reward: $R_{i}$=R\;
Update agent: $\theta_{\pi} \gets \theta_{\pi} +\alpha\sum_{i=1}^{|B^k|}R_{i} \nabla_{\theta_{\pi}}\log\pi$
}
$\theta_{\pi}'=\tau\theta_{\pi}+(1-\tau)\theta_{\pi}'$\;
Update the classifier $f_{b}$\;
}
\caption{PR REINFORCE}
\end{algorithm}

\section{Experiment}
Our experiment is designed to demonstrate that our proposed methodologies can train an instance selector more efficiently.

We tuned our model using three-fold cross validation on  the training set. For the parameters of the instance selector, we set the dimension of entity embedding as 50, the learning rate as 0.01. The delay coefficient $\tau$ is 0.005.
For the parameters of the relation classifier,  we follow the settings that are described in \cite{BGRU2016}.

The comparison is done in rule-based reinforcement learning method, original reinforcement learning and method with no reinforcement learning which is the basic relation classifier trained on original DS dataset. We use the last as the baseline.

\subsection{Dataset}
A widely used DS dataset, which is developed by ~\cite{riedel2010}, is used as the original dataset to be selected.
The dataset is generated by aligning Freebase with New York Times corpus.

\subsection{Metric and Performance Comparison}
We compare the data selection model performance by the final performance of the basic model trained on newly constructed dataset selected by different models. We use the precision/recall curves as the main metric. Fig.3 presents this comparison. PR REINFORCE constructs cleaned DS dataset with less noisy data compared with the original REINFORCE so that the BGRU+2ATT classifier can reach better performance.

%As for training efficiency, we measure the time we spend in the training procedure of PR
%REINFORCE as well as original REINFORCE. We train both models on a same device (i.e., NVIDIA GeForce GTX 1080 Ti). The time spend by this two model is 10.58 hours  and 14.35 hours respectively. PR REINFORCE gains 26.3\% increase of efficiency  compared with original REINFORCE.
\begin{figure}
\includegraphics[height=180pt, width=\columnwidth]{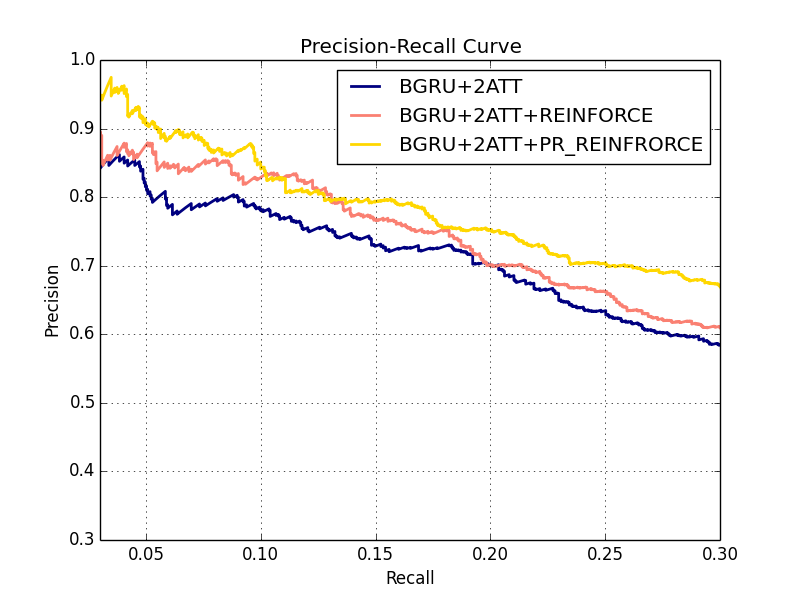}
\caption{Precision/Recall Curves}
\end{figure}

\section{Conclusions}
In this paper, we develop a posterior regularized REINFORCE methodology to alleviate the wrong label problem in distant supervision. Our model makes full use of the hand-crafted domain-specific rules in the trial and error search during the training process of REINFORCE method for DS dataset selection. The experiment results show that PR REINFORCE outperforms the original REINFORCE. Moreover, PR REINFORCE greatly improves  the efficiency of the REINFORCE training.

%\end{document}  % This is where a 'short' article might terminate

\section*{Acknowledgments}
This work has been supported in part by NSFC (No.61751209, U1611461), 973 program (No. 2015CB352302),
Hikvision-Zhejiang University Joint Research Center, Chinese
Knowledge Center of Engineering Science and Technology
(CKCEST), Engineering Research Center of Digital
Library, Ministry of Education. Xiang Ren's research
has been supported in part by National Science Foundation
SMA 18-29268.

\bibliography{naaclhlt2019}
\bibliographystyle{acl_natbib}

\end{document}